\documentclass[10pt,journal,compsoc]{IEEEtran}

\ifCLASSOPTIONcompsoc
  \usepackage[nocompress]{cite}
\else
  \usepackage{cite}
\fi
\usepackage[dvips]{graphicx}

\usepackage[linesnumbered,ruled,vlined]{algorithm2e}

\usepackage{url}

\usepackage{lineno}

\ifCLASSINFOpdf
\else
\fi
\usepackage[cmex10]{amsmath}
\usepackage{amssymb}
\usepackage{amsthm}
\usepackage{eqnarray}
\newtheorem{theorem}{Theorem}
\theoremstyle{definition}\newtheorem{definition}{Definition}

\DeclareMathOperator*{\argmin}{argmin}

\begin{document}
%
\title{A New Low-Rank Tensor Model for Video Completion}
%
%
%
%

\author{Wenrui~Hu,
        Dacheng~Tao,
        Wensheng~Zhang,
        Yuan~Xie,
        and ~Yehui~Yang
\IEEEcompsocitemizethanks{
\IEEEcompsocthanksitem  W. Hu,  W. Zhang, Y. Xie, and Y. Yang are with the Research Center of Precision Sensing and Control, Chinese Academy of Sciences, Beijing, 100190, China; E-mail: \{wenrui.hu, wensheng.zhang, yuan.xie, yehui.yang\}@ia.ac.cn
\IEEEcompsocthanksitem D. Tao is with the Center for Quantum Computation $\&$ Intelligent Systems and the Faculty of Engineering $\&$ Information Technology, University of Technology, Sydney, Australia; E-mail: dacheng.tao@uts.edu.au}
\thanks{}}

%
%

\markboth{}%
{Shell \MakeLowercase{\textit{et al.}}: Bare Advanced Demo of IEEEtran.cls for Journals}
%



\IEEEtitleabstractindextext{%
\begin{abstract}

In this paper, we propose a new low-rank tensor model based on the circulant algebra, namely, twist tensor nuclear norm or t-TNN for short. The twist tensor denotes a 3-way tensor representation to laterally store 2D data slices in order. On one hand, t-TNN convexly relaxes the tensor multi-rank of the twist tensor in the Fourier domain, which allows an efficient computation using FFT. On the other, t-TNN is equal to the nuclear norm of block circulant matricization of the twist tensor in the original domain, which extends the traditional matrix nuclear norm in a block circulant way. We test the t-TNN model on a video completion application that aims to fill missing values and the experiment results validate its effectiveness, especially when dealing with video recorded by a non-stationary panning camera. The block circulant matricization of the twist tensor can be transformed into a circulant block representation with nuclear norm invariance. This representation, after transformation, exploits the horizontal translation relationship between the frames in a video, and endows the t-TNN model with a more powerful ability to reconstruct panning videos than the existing state-of-the-art low-rank models.

\end{abstract}

}

\maketitle

\IEEEdisplaynontitleabstractindextext

%
\IEEEpeerreviewmaketitle


\ifCLASSOPTIONcompsoc
\IEEEraisesectionheading{\section{Introduction}\label{sec:introduction}}
\else
\section{Introduction}
\label{sec:introduction}
\fi

%
%
%
%
\IEEEPARstart{L}{ow-rank} tensor estimation (LRTE), which reveals the algebraic structure of multi-dimensional data (also referred to as tensors), is  a rapidly growing feature of research in many areas, such as computer vision \cite{goldfarb13}, signal processing \cite{rajwade13}, data mining \cite{kolda08} and machine learning \cite{signoretto14}. At the core of LRTE lies low-rank tensor decomposition \cite{kolda09}, and for tensors of order higher than 2, two decompositions are commonly used, i.e., CANDECOMP/PARAFAC(CP) \cite{carroll70} and Tucker decomposition \cite{tucker66}.

The CP model factorizes a tensor into a sum of rank-1 tensors, but it suffers from known computational and ill-posedness issues \cite{silva08}. The Tucker model, as an economical surrogate, extends the notion of matrix rank to rank-$N$ for an $N$-dimension tensor \cite{lathauwer00}, and then forces the unfolding matrices of the tensor along each mode (i.e., single dimension) to be low-rank using the matrix SVD-based factorization method. It is usually necessary to specify the rank of each unfolding matrix as a prior in the utilization of Tucker decomposition, which tends to over/under-estimate the truth \cite{chen14}. Moreover, Tucker decomposition suffers from local minima \cite{kolda09} as a result of non-convex optimization.

To overcome the above-mentioned drawbacks of the Tucker model, a convex relaxation technique, namely the sum of nuclear norms (SNN) of unfolding matrices, is provided by some authors \cite{gandy11, liu13, mu14}. SNN penalizes all unfolding matrices with the nuclear norm and serves as a tractable measure of rank-$N$ in practical settings. All modes being simultaneously low-rank might be strong assumption for a tensor, however. In consideration of this, a latent nuclear norm model (LNN) is introduced in \cite{tomioka13}. LNN only requires that the tensor be the sum of a set of component tensors, each of which is low-rank in the corresponding mode, and this strategy enables LNN to automatically detect the rank-deficient modes.

Note that neither SNN nor LNN exploits the correlations between modes. Furthermore, they try to model the tensor in the matrix SVD-based vector space, which results in loss of optimality in the representation. With these motivations, a tensor nuclear norm (TNN) is introduced into the LRTE problem for various tasks \cite{ely13, semerci14, zhang14}. The TNN model is based upon a new tensor decomposition scheme in \cite{braman10, martin13, kilmer13} which the authors refer to as tensor-SVD or t-SVD for short. t-SVD has a similar structure to the matrix SVD and models a tensor in the matrix space through a defined t-product operation \cite{kilmer13}. By transforming into the nuclear norm of block circulant representation, TNN simultaneously characterizes the low rank of a tensor along various modes.

In this paper, we propose a new low-rank tensor model, twist tensor nuclear norm (t-TNN), for 3D video completion, in which the video sequence recorded by a stationary or non-stationary camera is considered as a low-rank tensor (exactly or approximately). In the t-TNN model, we design a 3-way tensor representation named twist tensor which laterally stores 2D data slices in order; the twist tensor is then used to exploit the low-rank structure of data based on the t-SVD framework. By equivalizing the nuclear norm of the block circulant matricization of the twist tensor, t-TNN bridges the t-SVD based tensor nuclear norm and the traditional matrix nuclear norm (MNN) \cite{candes09}. This bridging enables t-TNN not only to exploit the correlations between all the modes simultaneously but also to take advantage of the low-rank prior along a certain mode which is rooted in some types of tensor data, e.g., video sequence over the time dimension, hyperspectral images via the wavelength variable, and face image samples through the number index.

In the video completion application, the t-TNN model is verified as being more effective in reconstructing texture and fine detail than the existing state-of-the-art low-rank models (including the generalized TNN (GTNN) in \cite{ely13, semerci14, zhang14}), especially when dealing with video recorded by a non-stationary panning camera. We interpret this phenomenon by transforming the block circulant matricization of the twist tensor into a circulant block representation with nuclear norm invariance. This representation exploits the horizontal translation relationship between frames in a video, which gives the t-TNN model a suitable low-rank description for panning videos without translation compensation.

The rest of this paper is organized as follows. Section \ref{sec:related_works} introduces related works and highlights the challenges of video completion. Section \ref{sec:notations} gives the preliminaries on tensors and the notations that will be used throughout the paper. Section \ref{sec:method} describes the proposed model and algorithm for the LRTE problem in detail. Experimental analysis and completion results are given in Section \ref{sec:experiment} to verify our method. Lastly, Section \ref{sec:conclusion} gives concluding remarks and future directions.

\section{Related Works of Video Completion}\label{sec:related_works}
\begin{figure}[htb]
\setlength{\abovecaptionskip}{0pt}  
\setlength{\belowcaptionskip}{0pt} 
\renewcommand{\figurename}{Figure}
\centering
\includegraphics[width=0.4\textwidth]{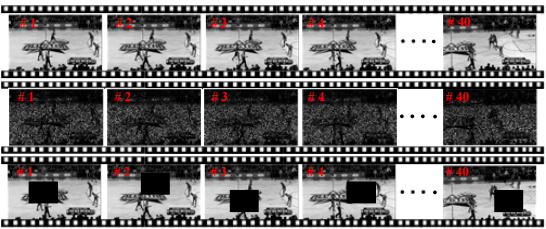}
\caption{A panning video sequence existing multiple moving objects. Missing pixel (middle panel) imputation and occlusion (bottom panel) correction can be formalized as the video completion problem. An important characteristic of video data is the temporal redundancy, and the effective utilization of this prior information can improve the performance of video completion.}
\label{fig:vision_completion}
\end{figure}

Video completion or inpainting is a computer vision technique to fill the missing values of video sequences in a seamless manner. It is critical for applications in video repairing, video editing, and movie postproduction, to name but a few \cite{wang14}. The missing values can be caused by many circumstances \cite{ji11}, e.g., natural noise in the video capture equipment, errors in data conversion/communication, occlusion by obstacles in the environment, and the segmenting or removal of interesting objects from videos. Fig. \ref{fig:vision_completion} illustrates the degradation of a video by insufficient sampling with a rate 0.3 (see the middle panel) and random occlusion with a $25 \times 25$ black block (see the bottom panel).

As the space-time equivalent of image completion \cite{zarif15}, video completion inherits and extends the solutions of the original 2D problem. For example, \cite{bertalmio01} utilizes the spatial partial differential equations (PDEs) based image inpainting method \cite{bertalmio00} to complete the video frame-by-frame, and in \cite{jia04}, Jia et. al. extend the image repairing method \cite{jia03} to the video case. Video completion also imposes a number of challenges beyond completion of the image, however - mainly temporal coherency and spatial complexity. As a result, the method in \cite{bertalmio01} often causes more abrupt on temporal edges than spatial edges, while \cite{jia04} involves a gamut of different techniques that make the process of video completion very complicated \cite{patwardhan07}. Considering that temporal information can significantly improve video completion, some works, e.g., \cite{matsushita05} and \cite{wexler07}, exploit the temporal redundancy at the cost of high computational burden. Most recently, low-rank based methods \cite{liu13, zhang14, xu15} have achieved good estimations on missing values by counting the global space-time information \cite{wang14}.

In this paper, we hold that the effectiveness of video completion largely depends on the effective utilization of the temporal redundancy between frames and the spatial relationships between entries. By exploiting the low-rank property of the video data in the twist tensor representation, our method is able to deal with a variety of challenging situations that arise in video completion, such as the correct reconstruction of dynamic textures, multiple moving objects and moving background \cite{newson14}. Here, we assume that the motion of the background is caused by a camera with pan motions parallel to the scene, which occurs in a wide range of real circumstances \cite{patwardhan07}.

\section{Notations and Preliminaries} \label{sec:notations}
In this section, we introduce the notations and give the basic definitions used in the rest of the paper. We use calligraphy letters for tensors, e.g., $\mathcal{X}$, upper case letters for matrices, e.g., $X$, bold lower case letters for vectors, e.g., $\mathbf{x}$, and lower case letters for the entries, e.g., $x_{ij}$. The Frobenius norm of a matrix $X$ is defined as $||X||_{F} := (\sum_{i,j}|x_{ij}|^{2})^{\frac{1}{2}}$. Let $X = U \Sigma V^{\mathrm{T}}$ be the SVD of $X$ and $\sigma_{i}(X)$ the $i$th largest singular value, then the MNN of $X$ is $||X||_{\ast} := \sum_{i}\sigma_{i}(X)$. The corresponding singular-value thresholding (SVT) operation with threshold $\tau$ is $\mathbf{\mathrm{D}}_{\tau}(X)=U\Sigma_{\tau}V^{\mathrm{T}}$, where $\Sigma_{\tau} = \mathrm{diag}\left\{\left(\sigma_{i}(X)-\tau \right)_{+}\right\}$ and $t_{+}$ is the positive part of $t$.

An $N$-way (or $N$-mode) tensor is a multi-linear structure in $\Re^{n_{1} \times n_{2} \times \ldots \times n_{N}}$. A slice of an tensor is a 2D section defined by fixing all but two indices, and a fiber is a 1D section defined by fixing all indices but one \cite{kolda09}. For a 3-way tensor $\mathcal{X}$, we will use the Matlab notation $\mathcal{X}(k, :, :)$, $\mathcal{X}(:, k, :)$ and $\mathcal{X}(:, :, k)$ to denote respectively the $k$th horizontal, lateral and frontal slices, $\mathcal{X}(:, i, j)$, $\mathcal{X}(i, :, j)$ and $\mathcal{X}(i, j, :)$ to denote the mode-1, mode-2 and mode-3 fibers, and $\mathcal{X}_{f} = \mathrm{fft}(\mathcal{X},[~],3)$ to denote the Fourier transform along the third dimension. In particular, $\mathcal{X}^{(k)}$ is used to represent ${\mathcal{X}}(:, :, k)$. The mode-$l$ unfolding ${\mathcal{X}}_{(l)} \in \Re^{n_{l} \times \prod_{l'\neq l} n_{l'}}$ is a matrix whose columns are mode-$l$ fibers \cite{kolda09}. The opposite operation ``fold'' of the unfolding is defined as $\mathrm{fold}_{l}({\mathcal{X}}_{(l)}) = {\mathcal{X}}$. The Frobenius norm of ${\mathcal{X}}$ is $||{\mathcal{X}}||_{F} :=  (\sum_{i,j,k}|x_{ijk}|^{2})^{\frac{1}{2}}$, and the $l_{1}$ norm of ${\mathcal{X}}$ is $||{\mathcal{X}}||_{1} := \sum_{i,j,k}|x_{ijk}|$.

To construction our tensor nuclear norm based on t-SVD, it is necessary to introduce five block-based operators, i.e., $\mathrm{bcirc}$, $\mathrm{bvec}$, $\mathrm{bvfold}$, $\mathrm{bdiag}$ and $\mathrm{bdfold}$ \cite{kilmer13}, in advance. For ${\mathcal{X}} \in \Re^{n_{1} \times n_{2} \times n_{3}}$ specially, the ${\mathcal{X}}^{(k)}$s can be used to form the block circulant matrix
\begin{equation}\label{fml:bcm}
\mathrm{bcirc}({\mathcal{X}}) :=
\left[
\begin{matrix}
 {\mathcal{X}}^{(1)}    &{\mathcal{X}}^{(n_{3})}  & \cdots         & {\mathcal{X}}^{(2)} \\
 {\mathcal{X}}^{(2)}    &{\mathcal{X}}^{(1)}      &  \cdots        & {\mathcal{X}}^{(3)} \\
 \vdots            &\ddots              & \ddots         & \vdots         \\
 {\mathcal{X}}^{(n_{3})}&{\mathcal{X}}^{(n_{3}-1)}& \cdots         & {\mathcal{X}}^{(1)}
\end{matrix}
\right],
\end{equation}
the block vectorizing and its opposite operation
\begin{equation}\label{fml:bvec_bvfold}
\mathrm{bvec}({\mathcal{X}}) :=
\left[
\begin{matrix}
 {\mathcal{X}}^{(1)} \\
 {\mathcal{X}}^{(2)} \\
 \vdots         \\
 {\mathcal{X}}^{(n_{3})}
\end{matrix}
\right],~~~
\mathrm{bvfold}(\mathrm{bvec}({\mathcal{X}})) = {\mathcal{X}},
\end{equation}
and the block diag matrix and its opposite operation
\begin{equation}
\label{fml:bdiag}
\mathrm{bdiag}({\mathcal{X}}) :=
\left[
\begin{matrix}
{\mathcal{X}}^{(1)} & & \\
 &\ddots & \\
 && {\mathcal{X}}^{(n_{3})}
\end{matrix}
\right],~~~
\mathrm{bdfold}(\mathrm{bdiag}({\mathcal{X}})) = {\mathcal{X}}.
\end{equation}
The t-product between two 3-way tensors can then be defined as follows \cite{kilmer13} :

\begin{definition}[\textbf{t-product}]
\label{def:t-prod}
Let ${\mathcal{X}}$ be $n_{1} \times n_{2} \times n_{3}$ and ${\mathcal{Y}}$ be $n_{2} \times n_{4} \times n_{3}$. The t-product ${\mathcal{X}}*{\mathcal{Y}}$ is an $n_{1} \times n_{4} \times n_{3}$ tensor
\begin{equation}
\label{fml:t-prod}
{\mathcal{M}} = {\mathcal{X}}*{\mathcal{Y}} = : \mathrm{bvfold}\{\mathrm{bcirc}({\mathcal{X}})\mathrm{bvec}({\mathcal{Y}})\}.
\end{equation}
\end{definition}

The t-product is analogous to the matrix multiplication except that the circular convolution replaces the multiplication operation between the elements, which are now mode-3 fibers \cite{zhang14}, as follows :
\begin{equation}
\label{fml:t-prod2}
{\mathcal{M}}(i,j,:) = \sum_{k=1}^{n_{2}}{\mathcal{X}}(i,k,:) \circ {\mathcal{Y}}(k,j,:),
\end{equation}
where $\circ$ denotes the circular convolution between two tubes. The t-product in the original domain corresponds to the matrix multiplication of the frontal slices in the Fourier domain, as follows :
\begin{equation}
\label{fml:t-prod3}
{\mathcal{M}}_{f}^{(k)} = {\mathcal{X}}_{f}^{(k)} {\mathcal{Y}}_{f}^{(k)},~ k = 1,\ldots,n_{3}.
\end{equation}

Next we define the related notions of the tensor transpose, identity tensor, orthogonal tensor and f-diagonal tensor \cite{kilmer13}.

\begin{definition}[\textbf{Tensor Transpose}]
\label{def:t-trans}
Let ${\mathcal{X}} \in \Re^{ n_{1} \times n_{2} \times n_{3} }$, the transpose tensor ${ \mathcal{X}}^{\mathrm{T}}$ is an $n_{2} \times n_{1} \times n_{3}$ tensor obtained by transposing each frontal slice of ${\mathcal{X}}$ and then reversing the order of the transposed frontal slices 2 through $n_3$.
\end{definition}

\begin{definition}[\textbf{Identity Tensor}]
\label{def:t-I}
The identity tensor ${\mathcal{I}} \in \Re^{n_{1} \times n_{1} \times n_{3}}$ is a tensor whose first frontal slice is the $n_{1} \times n_{1}$ identity matrix and all other frontal slices are zero.
\end{definition}

\begin{definition}[\textbf{Orthogonal Tensor}]
\label{def:orth-tensor}
A tensor ${\mathcal{Q}} \in \Re^{n_{1} \times n_{1} \times n_{3}}$ is orthogonal if
\begin{equation}
{\mathcal{Q}}^{\mathrm{T}}*{\mathcal{Q}} = {\mathcal{Q}}*{\mathcal{Q}}^{\mathrm{T}} = {\mathcal{I}},
\end{equation}
where $*$ is the t-product.
\end{definition}

\begin{definition}[\textbf{f-diagonal Tensor}]
\label{def:f-diag}
A tensor is called f-diagonal if each of its frontal slices is diagonal matrix. The t-production of two f-diagonal tensors with the same size $n_{1} \times n_{2} \times n_{3}$, i.e., $\mathcal{M}=\mathcal{X}*\mathcal{Y}$, is also an $n_{1} \times n_{2} \times n_{3}$ f-diagonal tensor, and its diagonal tube fibers are
\begin{equation}
\label{fml:tprod_fdiag}
\mathcal{M}(i,i,:) = \mathcal{X}(i,i,:) \circ \mathcal{Y}(i,i,:),~ i = 1, \ldots, \mathrm{min}(n_{1}, n_{2}).
\end{equation}
\end{definition}

\section{Method}\label{sec:method}
In this section we describe the proposed method in detail. In Section \ref{subsec:gtnn}, we introduce t-SVD and the corresponding tensor multi-rank, then GTNN is provided for the purpose of convex relaxation. In Section \ref{subsec:ttnn}, a twist tensor is first defined, which leads to our new low-rank tensor model t-TNN. We also investigate the relationship between t-TNN and the traditional MNN of mode-3 unfolding and compare t-TNN with GTNN by transforming t-TNN into the circulant block representation. Section \ref{subsec:method_tc} presents the optimization algorithm for the t-TNN based tensor completion.

\subsection{Generalized tensor nuclear norm (GTNN)}\label{subsec:gtnn}
\begin{figure}[htb]
\setlength{\abovecaptionskip}{0pt}  
\setlength{\belowcaptionskip}{0pt} 
\renewcommand{\figurename}{Figure}
\centering
\includegraphics[width=0.3\textwidth]{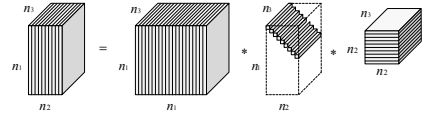}
\caption{The t-SVD of an $n_{1} \times n_{2} \times n_{3}$ tensor.}
\label{fig:tSVD}
\end{figure}

For a 3-way tensor ${\mathcal{X}} \in {\Re}^{n_{1} \times n_{2} \times n_{3}}$, the t-SVD of ${\mathcal{X}} $ is given by
\begin{equation}\label{fml:t-svd}
{\mathcal{X}}={\mathcal{U}}*{\mathcal{S}}*{\mathcal{V}}^{\mathrm{T}},
\end{equation}
where $\mathcal{U}$ and $\mathcal{V}$ are orthogonal tensors of size $n_{1} \times n_{1} \times n_{3}$ and $n_{2} \times n_{2} \times n_{3}$ respectively. $\mathcal{S}$ is an f-diagonal tensor of size $n_{1} \times n_{2} \times n_{3}$, and $*$ denotes the t-product. Fig. \ref{fig:tSVD} illustrates the decomposition. As demonstrated in Eq. (\ref{fml:t-prod3}), the t-production can be computed efficiently in the Fourier domain, which leads Alg. \ref{alg:tSVD} to obtain the t-SVD \cite{kilmer13}.

\begin{algorithm}[]\label{alg:tSVD}
\SetAlgoLined
\caption{t-SVD}
\KwIn{~~~${\cal X} \in \Re^{n_{1} \times n_{2} \times n_{3}}$.\\}
\KwOut{~${\cal U}$,~${\cal S}$,~${\cal V}$.\\}
\BlankLine
${\cal X}_{f} = \mathrm{fft}({\cal X},[~],3)$.\\
\For{$k = 1 : n_{3}$}
{
$[U,\Sigma,V] = \mathrm{SVD}({\cal X}_{f}^{(k)})$.\\
${\cal U}_{f}^{(k)} = U$,~${\cal S}_{f}^{(k)} = \Sigma$,~${\cal V}_{f}^{(k)} = V$.\\
}
${\cal U} = \mathrm{ifft}({\cal U}_{f},[~],3)$,~${\cal S} = \mathrm{ifft}({\cal S}_{f},[~],3)$,~ ${\cal V} = \mathrm{ifft}({\cal V}_{f},[~],3)$.\\
\BlankLine
\textbf{Return} $\cal U$, $\cal S$, $\cal V$\;
\end{algorithm}

Resorting to t-SVD, we can define the tensor multi-rank as follows \cite{zhang14, semerci14, kilmer13} :
\begin{definition}[\textbf{Tensor multi-rank}]
\label{def:multi-rank}
The multi-rank of ${\mathcal{X}} \in \Re^{n_{1} \times n_{2} \times n_{3}}$ is a vector $\mathbf{r} \in \Re^{n_{3} \times 1}$ with the $i_{th}$ element equal to the rank of the $i_{th}$ frontal slice of ${\mathcal{X}}_{f}$.
\end{definition}
Then the GTNN is given as
\begin{equation}
\label{fml:gtnn}
||{\mathcal{X}}||_{\circledast} := \sum_{k=1}^{n_{3}}\sum_{i=1}^{\mathrm{min}(n_{1},n_{2})}|{ \mathcal{S}}_{f}(i,i,k)|,
\end{equation}
which is proven to be a valid norm and the tightest convex relaxation to $\ell_{1}$ norm of the tensor multi-rank in  \cite{semerci14,zhang14}. From Alg. \ref{alg:tSVD}, it can be seen that
\begin{equation}
\label{fml:mf_svd}
\mathrm{bdiag}(\mathcal{X}_{f})=\mathrm{bdiag}(\mathcal{U}_{f})\mathrm{bdiag}(\mathcal{S}_{f})\mathrm{bdiag}(\mathcal{V}_{f})^{\mathrm{T}}.
\end{equation}
Due to the unitary invariance of MNN, we have
\begin{equation}
\label{fml:mf_nuclear_norm}
||\mathrm{bdiag}(\mathcal{X}_{f})||_{*} = ||\mathrm{bdiag}(\mathcal{S}_{f})||_{*} = ||{\mathcal{X}}||_{\circledast},
\end{equation}
and since block circulant matrixes can be block diagonalized by using the Fourier transform, there is
\begin{equation}
\label{fml:mf-bcirc}
\begin{aligned}
||\mathrm{bdiag}(\mathcal{X}_{f})||_{*} & = ||(F_{n_{3}}\otimes I_{n_{1}}) \mathrm{bcirc}({\mathcal{X}}) (F_{n_{3}}^{*}\otimes I_{n_{2}})||_{*}\\
                                         &=|| \mathrm{bcirc}({\mathcal{X}})||_{*}.
\end{aligned}
\end{equation}
Finally, we obtain
\begin{equation}
\label{fml:gtnn_birc}
||{\mathcal{X}}||_{\circledast} = || \mathrm{bcirc}({\mathcal{X}})||_{*}.
\end{equation}

The equivalence in Eq. (\ref{fml:gtnn_birc}) endows the GTNN with interpretability in the original domain. We can see that the block circulant representation of ${\mathcal{X}}$ preserves the spacial relationship between entries, and $|| \mathrm{bcirc}({\mathcal{X}})||_{*}$ measures the rank of $\mathrm{bcirc}({\mathcal{X}})$ by comparing every row and every column of frontal slices over the third dimension (especially the time dimension for video data), which exploits the spatial-temporal information of a tensor deeper than the monotonous MNN of certain unfolding.

\subsection{Twist tensor nuclear norm (t-TNN)} \label{subsec:ttnn}
Before introducing our new tensor nuclear norm, we need to define the twist tensor as follows :

\begin{figure}[htb]
\setlength{\abovecaptionskip}{0pt}  
\setlength{\belowcaptionskip}{0pt} 
\renewcommand{\figurename}{Figure}
\centering
\includegraphics[width=0.18\textwidth]{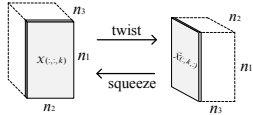}
\caption{The twist and squeeze operations.}
\label{fig:twist_squeeze}
\end{figure}

\begin{definition}[\textbf{Tensor Twist and Squeeze}]
\label{def:squeeze-tensor}
Let ${\mathcal{X}} \in \Re^{n_{1} \times n_{2} \times n_{3}}$, then the twist tensor ${\mathcal{Y}} = \mathord{\buildrel{\lower2pt\hbox{$\scriptscriptstyle\rightarrow$}} \over {\mathcal{X}}}$ is an $n_{1} \times n_{3} \times n_{2}$ tensor whose lateral slice ${\mathcal{Y}}(:,k,:) = \mathrm{twist}({\mathcal{X}}^{(k)})$. Correspondingly, the squeeze tensor of ${\mathcal{Y}}$, i.e., ${\mathcal{X}} = \mathord{\buildrel{\lower2pt\hbox{$\scriptscriptstyle\leftarrow$}} \over {\mathcal{Y}}}$, can be obtained by the reverse process, i.e., ${\mathcal{X}}^{(k)} = \mathrm{squeeze}({\mathcal{Y}}(:,k,:))$. Fig. \ref{fig:twist_squeeze} illustrates the twist and squeeze operations.
\end{definition}

The t-TNN based on the t-SVD framework is then
\begin{equation}
\label{fml:ttnn_def}
||{{\mathcal{X}}}||_{\vec{\circledast}} := ||\mathord{\buildrel{\lower2pt\hbox{$\scriptscriptstyle\rightarrow$}} \over {\mathcal{X}}}||_{\circledast} = || \mathrm{bcirc}(\mathord{\buildrel{\lower2pt\hbox{$\scriptscriptstyle\rightarrow$}} \over {\mathcal{X}}})||_{*}.
\end{equation}

\begin{figure}[htb]
\setlength{\abovecaptionskip}{0pt}  
\setlength{\belowcaptionskip}{0pt} 
\renewcommand{\figurename}{Figure}
\centering
\includegraphics[width=0.46\textwidth]{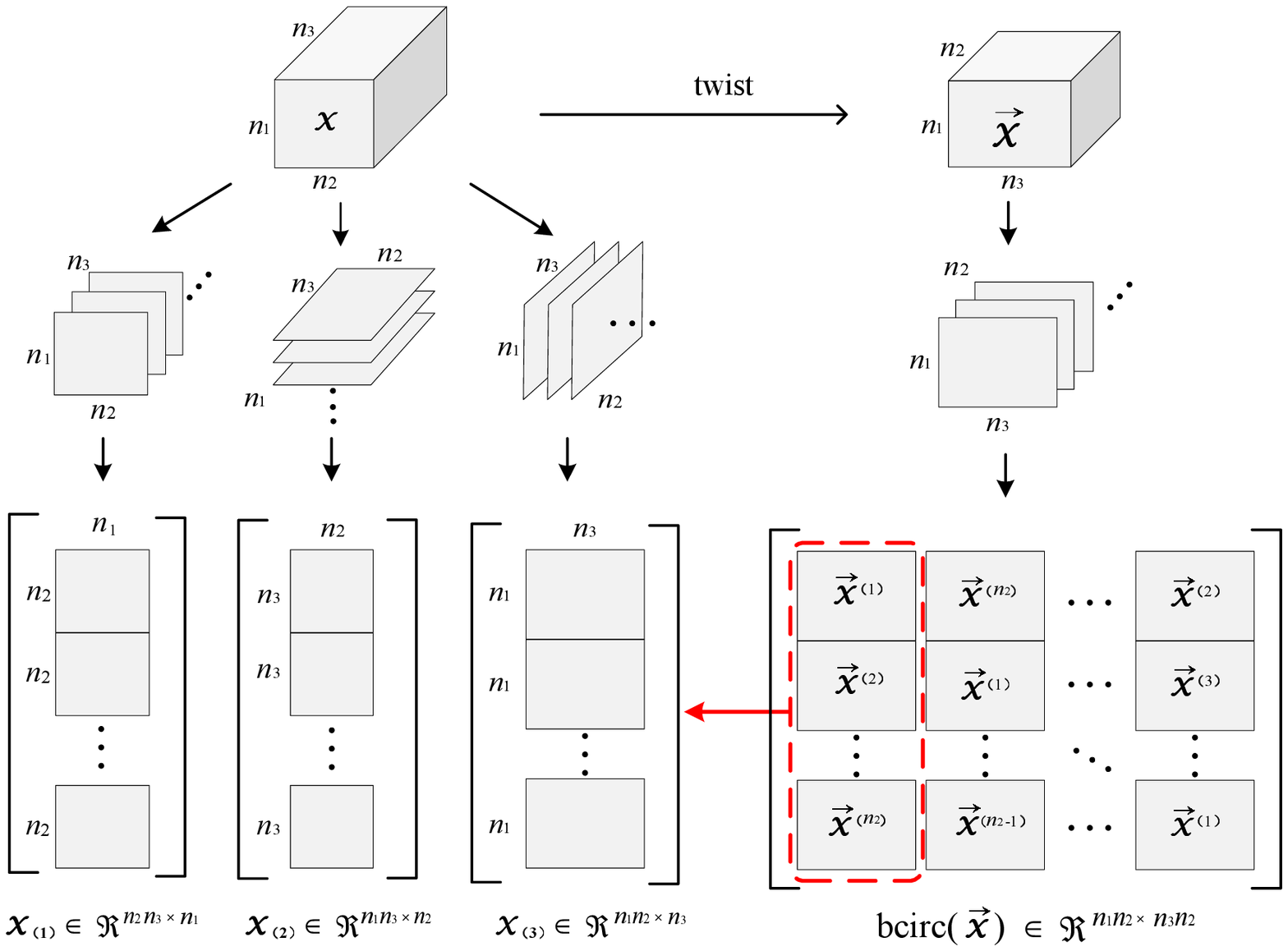}
\caption{Mode-1, mode-2 and mode-3 unfoldings of a 3-way tensor ${\cal X}$ and the block circulant matricization of the twist tensor $\mathord{\buildrel{\lower2pt\hbox{$\scriptscriptstyle\rightarrow$}} \over {\cal X}}$. The unfolding operation corresponds to aligning the corresponding slices for each mode next to each other. It can be seen that the first block column of $\mathrm{bcirc}(\mathord{\buildrel{\lower2pt\hbox{$\scriptscriptstyle\rightarrow$}} \over {\cal X}})$, i.e., $\mathrm{bvec}(\mathord{\buildrel{\lower2pt\hbox{$\scriptscriptstyle\rightarrow$}} \over {\cal X}})$, is equal to the mode-3 unfolding of ${\cal X}$.}
\label{fig:ntsvd_model}
\end{figure}

For a deeper insight into t-TNN, we  mine the relationship between t-TNN and the traditional MNN, which is illustrated in Fig. \ref{fig:ntsvd_model}. Consider a 3D video data ${\mathcal{X}}$ with size $n_1 \times n_2 \times n_3$ ($\mathrm{length} \times \mathrm{width} \times \mathrm{frames}$). The mode-3 unfolding of ${\cal X}$ can be obtained by vectorizing each frame then aligning them in time order. Due to the content continuity, i.e., the pixels at the same location in consecutive frames tend to change small, $||{\mathcal{X}}_{(3)}||_{*}$ can effectively exploit the temporal redundancy between frames. However, the vectorization of frames ignores the spatial information of pixels within one frame. By extending ${\mathcal{X}}_{(3)}$ in a block circulant way, $\mathrm{bcirc}(\mathord{\buildrel{\lower2pt\hbox{$\scriptscriptstyle\rightarrow$}} \over {\mathcal{X}}})$ contains a latent spatial feature for each frame along the row direction.

\begin{figure}[htb]
\setlength{\abovecaptionskip}{0pt}  
\setlength{\belowcaptionskip}{0pt} 
\renewcommand{\figurename}{Figure}
\centering
\includegraphics[width=0.35\textwidth]{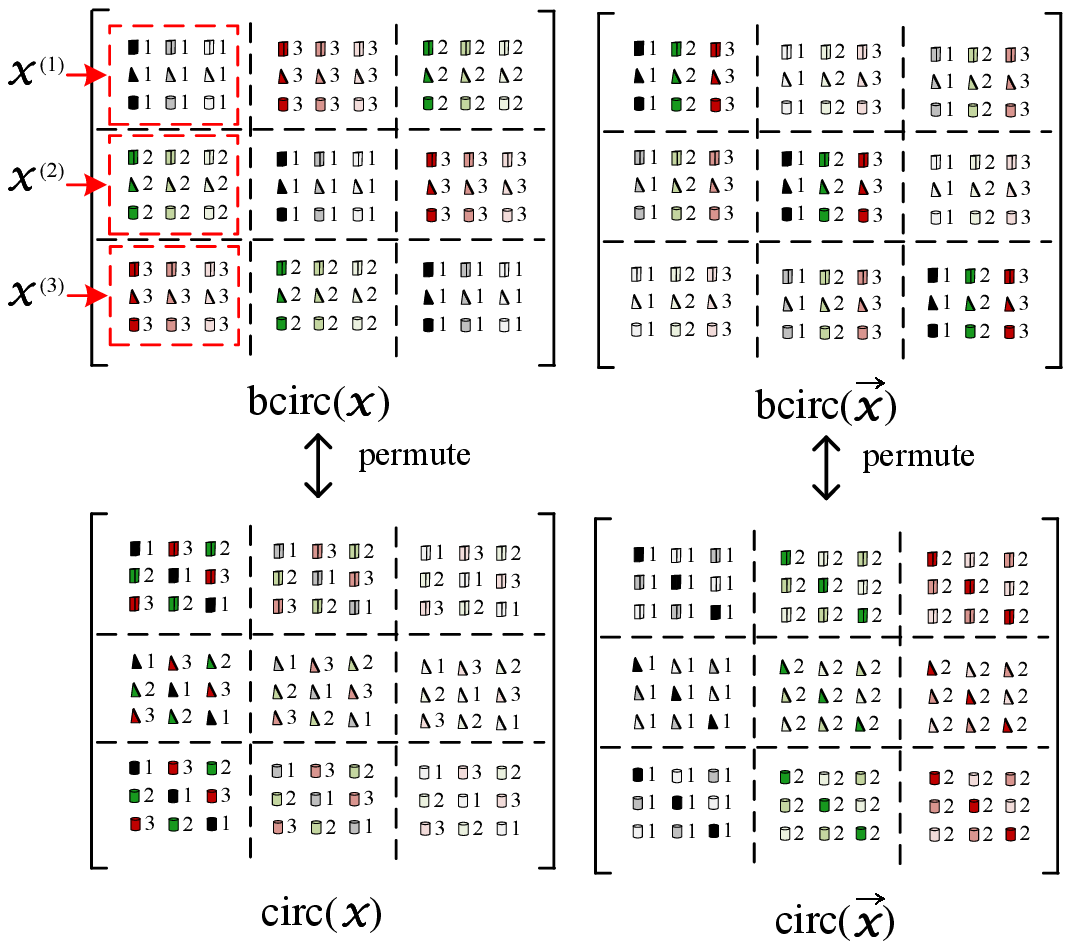}
\caption{Circulant block representation of $\mathcal{X}$ and $\mathord{\buildrel{\lower2pt\hbox{$\scriptscriptstyle\rightarrow$}} \over {\cal X}}$ by permuting $\mathrm{bcirc}(\mathcal{X})$ and $\mathrm{bcirc}(\mathord{\buildrel{\lower2pt\hbox{$\scriptscriptstyle\rightarrow$}} \over {\cal X}})$.}
\label{fig:circ_illustration}
\end{figure}

Next, we analyze the t-TNN model from the perspective of circulant block representation\cite{gleich13}. For a tensor ${\mathcal{X}}$ with size $m \times n \times k$, the circulant block matricization of ${\mathcal{X}}$ is defined as follows :
\begin{equation}\label{fml:circ}
\mathrm{circ}(\mathcal{X}) :=
\left[
\begin{matrix}
 \mathrm{circ}({\mathcal{X}}_{11:})    & \mathrm{circ}({\mathcal{X}}_{12:})    & \cdots         & \mathrm{circ}({\mathcal{X}}_{1n:}) \\
 \mathrm{circ}({\mathcal{X}}_{21:})    & \mathrm{circ}({\mathcal{X}}_{22:})    &  \cdots        & \mathrm{circ}({\mathcal{X}}_{2n:}) \\
 \vdots                             &\ddots                            & \ddots         & \vdots         \\
 \mathrm{circ}({\mathcal{X}}_{m1:})& \mathrm{circ}({\mathcal{X}}_{m2:}) & \cdots        & \mathrm{circ}({\mathcal{X}}_{mn:})
\end{matrix}
\right],
\end{equation}
where ${\mathcal{X}}_{ij:} = {\mathcal{X}}(i,j,:)$ is a mode-3 fiber which belongs to a length-$k$ vector space or $\mathbb{K}_{k}$, and $\mathrm{circ}({\mathcal{X}}_{ij:})$ constructs a $\mathbb{K}_{k}$-circulant module or block with ${\mathcal{X}}_{ij:}$. By permutation, there exists a relationship between $\mathrm{circ}(\mathcal{X})$ and $\mathrm{bcirc}(\mathcal{X})$ as follows :
\begin{equation}
\label{fml:circ_bcirc}
\mathrm{circ}(\mathcal{X}) = P_{1}\mathrm{bcirc}(\mathcal{X})P_{2},
\end{equation}
where $P_{i}$, $i = 1,2$, denote so-called stride permutations \cite{granata92}. Fig. \ref{fig:circ_illustration} illustrates the transformation of both $\mathrm{bcirc}(\mathcal{X})$ and $\mathrm{bcirc}(\mathord{\buildrel{\lower2pt\hbox{$\scriptscriptstyle\rightarrow$}} \over {\mathcal{X}}})$ with a $3 \times 3 \times 3$ tensor as an example. Because of the permutation invariance, we have
\begin{equation}
\label{fml:equival_permu}
||\mathrm{circ}({\mathcal{X}})||_{*} = ||\mathrm{bcirc}({\mathcal{X}})||_{*}.
\end{equation}

Considering a 3D video data ${\mathcal{X}}$, ${\mathcal{X}}_{ij:}$ corresponds to a time tube at position $(i,j)$, while $\mathord{\buildrel{\lower2pt\hbox{$\scriptscriptstyle\rightarrow$}} \over { \mathcal{X}}}_{ik:}$ is the $i$th row of the $k$th frame. Compared to $\mathrm{circ}({\mathcal{X}})$, $\mathrm{circ}(\mathord{\buildrel{\lower2pt\hbox{$\scriptscriptstyle\rightarrow$}} \over {\cal X}})$ is more suitable for describing a scene with a global panning motion, because $||\mathrm{circ}(\mathord{\buildrel{\lower2pt\hbox{$\scriptscriptstyle\rightarrow$}} \over { \mathcal{X}}})||_{*}$ measures the extent of the change in the rows across frames in a circulant way. In other words, when a video sequence is recorded by a camera in panning motion, there exists the horizontal translation for pixels over time, and $\mathrm{circ}(\mathord{\buildrel{\lower2pt\hbox{$\scriptscriptstyle\rightarrow$}} \over {\mathcal{X}}})$ exploits this linear relationship between frames.

\subsection{Tensor completion from missing values}\label{subsec:method_tc}
The task of tensor completion is to recover the latent tensor ${\mathcal{M}}$ from missing values, which can be addressed by solving the following convex optimization problem
\begin{equation}\label{fml:obj_ntsvd_tc}
\min_{{\mathcal{X}}\in {\mathcal{T}}} ||{{\mathcal{X}}}||_{\vec{\circledast}},~~s.t.~~\mathcal{P}_{\Omega}({\mathcal{X}}) = \mathcal{P}_{\Omega}({\mathcal{M}}),
\end{equation}
where $\mathcal{P}_{\Omega}$ is the orthogonal projector onto the span of tensors vanishing outside of $\Omega$, namely,
\begin{equation}
\label{fml:projection}
{\mathcal{P}}_{\Omega}(\mathcal{X})(i,j,k)=
\begin{cases}
0, & \text{${\Omega}_{ijk}=0$} \\
x_{ijk}, & \text{${\Omega}_{ijk}=1$}
\end{cases}
\end{equation}
and $\mathcal{P}_{{\Omega}^{\bot}}$ is the complementary projection, i.e., $\mathcal{P}_{\Omega}(\mathcal{X})+\mathcal{P}_{{\Omega}^{\bot}}(\mathcal{X})=\mathcal{X}$.

We adopt the alternating direction method of multipliers (ADMM)\cite{boyd10} to solve problem (\ref{fml:obj_ntsvd_tc}). By introducing a new tensor variable ${\mathcal{Y}} = {\mathcal{X}}$, we obtain the following objective function
\begin{equation}
\label{fml:unconstrained_obj_ntsvd_tc}
\begin{aligned}
E({\mathcal{X}},{\mathcal{Y}},{\mathcal{W}}) = & ||{\mathcal{Y}}||_{\vec{\circledast}} + \mathbf{1}_{{\mathcal{X}}_{\Omega}={ \mathcal{M}}_{\Omega}} +<{\mathcal{W}},{\mathcal{X}} - { \mathcal{Y}}> \\
& + \frac{\rho}{2} ||{\mathcal{X}}-{\mathcal{Y}}||_{F}^{2},
\end{aligned}
\end{equation}
where $\mathbf{1}$ denotes the indicator function. According to the framework of ADMM, we can iteratively update ${\mathcal{X}}$, ${\mathcal{Y}}$ and ${\mathcal{W}}$, as follows :
\begin{align}\label{fml:obj_ntsvd_tc_x}
{ \mathcal{X}} = & \argmin_{{\mathcal{X}}:{\mathcal{X}}_{\Omega}={ \mathcal{M}}_{\Omega}} ||{\mathcal{X}} - ({\mathcal{Y}} - \frac{1}{\rho}{\mathcal{W}})||_{F}^{2},\\
{\mathcal{Y}} = & ~\argmin_{{\mathcal{Y}}} ~ ||{\mathcal{Y}}||_{\vec{\circledast}} + \frac{\rho}{2}||{\mathcal{Y}} - ({\mathcal{X}} + \frac{1}{\rho}{ \mathcal{W}})||_{F}^{2}\label{fml:obj_ntsvd_tc_y},\\
{\mathcal{W}} = &  ~ {\mathcal{W}} + \rho({\mathcal{X}} - {\mathcal{Y}})\label{fml:obj_ntsvd_tc_w},
\end{align}
where Eq. (\ref{fml:obj_ntsvd_tc_x}) is a least-square projection onto the constraint and its solution is
\begin{equation}
\label{fml:obj_ntsvd_tc_x_solution}
\mathcal{X} = \mathcal{P}_{\Omega}(\mathcal{M}) + \mathcal{P}_{{\Omega}^{\bot}}({\mathcal{Y}} - \frac{1}{\rho}{ \mathcal{W}}).
\end{equation}

We can obtain the solution for Eq. (\ref{fml:obj_ntsvd_tc_y}) through the following theorem

\begin{theorem}
\label{theory:tsvc}
For $\tau > 0$ and $\mathcal{Y}, \mathcal{Z} \in \Re^{n_{1} \times n_{2} \times n_{3}}$, the twist tensor of the globally optimal solution to the following problem
\begin{equation}\label{fml:ntsvd}
\min_{{ \mathcal{Y}}} ~ \tau||{\mathcal{Y}}||_{\vec{\circledast}} + \frac{1}{2}||{\mathcal{Y}} - \mathcal{Z}||_{F}^{2}
\end{equation}
is given by the tensor singular value convoluting (tSVC)
\begin{equation}\label{fml:tsvc}
\mathord{\buildrel{\lower2pt\hbox{$\scriptscriptstyle\rightarrow$}} \over {\mathcal{Y}}} = \mathcal{C}_{\tau}(\mathord{\buildrel{\lower2pt\hbox{$\scriptscriptstyle\rightarrow$}} \over { \mathcal{Z}}} )=\mathcal{U}*\mathcal{C}_{\tau}(\mathcal{S}) *\mathcal{V}^{\mathrm{T}},
\end{equation}
where $\mathord{\buildrel{\lower2pt\hbox{$\scriptscriptstyle\rightarrow$}} \over { \mathcal{Z}}}=\mathcal{U}*\mathcal{S}*\mathcal{V}^{\mathrm{T}}$ and $\mathcal{C}_{\tau}(\mathcal{S}) = \mathcal{S}*\mathcal{J}$, herein, $\mathcal{J}$ is an $n_{1} \times n_{3} \times n_{2}$ f-diagonal tensor whose diagonal element in the Fourier domain is ${\mathcal{J}}_{f}(i,i,j) = (1 - \frac{\tau}{{ \mathcal{S}}_{f}^{(j)}(i,i)})_{+}$.
\end{theorem}

\begin{algorithm}[]\label{alg:ttnn_tc}
\SetAlgoLined
\caption{t-TNN based tensor completion}
\KwIn{~~~Observation data $\mathcal{M}$, projector $P_{\Omega}$\\}
\KwOut{~Completion tensor $\mathcal{X}$.\\}
\BlankLine
\textbf{Initialize:}~$\rho^{0}>0$,~$\eta > 1$,~$k = 0$,\\~~~~~~~~~~~$\mathcal{X}= P_{\Omega}({\cal M})$,~$\mathcal{Y}=\mathcal{W}=0$;\\
\While{$||{\cal X} - {\cal Y}||_{F}/||X||_{F} > tol$ ~and~$k < K$}
{
$\mathcal{X}^{k+1} = \mathcal{P}_{\Omega}(\mathcal{M}^{k}) + \mathcal{P}_{{\Omega}^{\perp}}({\cal Y}^{k} - \frac{1}{{\rho}^{k}}{\cal W}^{k})$;\\
$\tau = \frac{1}{{\rho}^{k}},~ \mathcal{Z} = \mathcal{X}^{k+1}+\tau{\cal W}^{k}$;\\
$\mathord{\buildrel{\lower2pt\hbox{$\scriptscriptstyle\rightarrow$}} \over {\cal Z}}_{f} = \mathrm{fft}(\mathord{\buildrel{\lower2pt\hbox{$\scriptscriptstyle\rightarrow$}} \over {\cal Z}},[~],3)$;\\
\For{$j = 1 : n_{2}$}
{
$[\mathcal{U}^{(j)}_{f},\mathcal{S}^{(j)}_{f},\mathcal{V}^{(j)}_{f}] = \mathrm{SVD}(\mathord{\buildrel{\lower2pt\hbox{$\scriptscriptstyle\rightarrow$}} \over {\cal Z}}^{(j)}_{f})$;\\
${\cal J}_{f}^{(j)} = \mathrm{diag}\{(1 - \frac{\tau}{{\cal S}_{f}^{(j)}(i,i)})_{+}\}$;\\
$\mathcal{S}^{(j)}_{f,\tau} = \mathcal{S}^{(j)}_{f}{\cal J}_{f}^{(j)}$;\\
$\mathcal{H}^{(j)}_{f} = \mathcal{U}_{f}^{(j)}\mathcal{S}_{f,\tau}^{(j)}\mathcal{V}_{f}^{(j)^{\mathrm{T}}}$;\\
}
$\mathcal{H} = \mathrm{ifft}(\mathcal{H}_{f},[~],3)$,~$\mathcal{Y}^{k+1} = \mathord{\buildrel{\lower2pt\hbox{$\scriptscriptstyle\leftarrow$}} \over {\cal H}}$;\\
${\cal W}^{k+1} =  ~ {\cal W}^{k} + {\rho}^{k}({\cal X}^{k+1} - {\cal Y}^{k+1})$;\\
${\rho}^{k+1} = \eta{\rho}^{k},~ k = k +1$
}
\BlankLine
\textbf{Return} Tensor $\cal X$;
\end{algorithm}

The proof of Theorem \ref{theory:tsvc} can be found in the Appendix, and the optimization procedure of our t-TNN based tensor completion is described in Alg. \ref{alg:ttnn_tc}. The convergence of ADMM has been proved in \cite{boyd10}, and the computational bottleneck of Alg. \ref{alg:ttnn_tc} lies in computing the 3D FFT and 3D inverse FFT of an $n_{1} \times n_{3} \times n_{2}$ tensor and $n_2$ SVDs of $n_{1} \times n_{3}$ matrices in the Fourier domain. There is no need to run over all $n_{2}$ SVDs because of the conjugate symmetry in the Fourier domain. For example, if $n_{2}$ is even, we can run the SVDs for $j = 1, \ldots, \frac{n_{2}}{2}+1$, and then populate the remaining $\frac{n_{2}}{2}-1$ SVDs for $j = 2,\ldots,\frac{n_{2}}{2}$ as follows:
\begin{align}\label{fml:svd_conj}
\mathcal{U}_{f}^{(n_{2}-j+2)} & = \mathrm{conj}(\mathcal{U}_{f}^{(j)}),\\
\mathcal{S}_{f}^{(n_{2}-j+2)} & = \mathcal{S}_{f}^{(j)},\\
\mathcal{V}_{f}^{(n_{2}-j+2)} &= \mathrm{conj}(\mathcal{V}_{f}^{(j)}).
\end{align}

Since with most situations in video data we have $n_{1},n_{2} > n_{3}$ and $\mathrm{log}(n_{2})<n_{1},n_{3}$, the computation at each iteration will take ${ \mathcal{O}}(2n_{1}n_{2}n_{3}\mathrm{log}(n_{2})+\frac{1}{2}n_{1}n_{2}n_{3}^{2})\approx { \mathcal{O}}(n_{1}n_{2}n_{3}^{2})$ (without considering the parallel computing for each SVD) compared to ${\mathcal{O}}(\mathrm{min}\{n_{1}^{2}n_{2},n_{1}n_{2}^{2}\}n_{3})$ using GTNN and ${\mathcal{O}}(n_{1}n_{2}n_{3}^{2})$ using MNN for the mode-3 unfolding.

\section{Experiments}\label{sec:experiment}
In this section we compare our t-TNN model with five other tensor-based or matrix-based models for real video completion: GTNN \cite{zhang14}, SNN \cite{liu13}, LNN \cite{tomioka13}, TMac \cite{xu15}, and MNN \cite{candes09}. All the completion methods but TMac are implemented in the ADMM framework, and their parameter settings are empirically determined to give the best performance and are fixed in all tests. For TMac, the completion results are generated from the source codes released by their authors\footnote{\url{http://www.math.ucla.edu/~wotaoyin/papers/tmac.html}.}. Nine videos datasets recorded by stationary camera or non-stationary camera are used to verify the effectiveness of the proposed model.

\begin{figure*}[htb]
\setlength{\abovecaptionskip}{0pt}  
\setlength{\belowcaptionskip}{0pt} 
\renewcommand{\figurename}{Figure}
\centering
\includegraphics[width=0.95\textwidth]{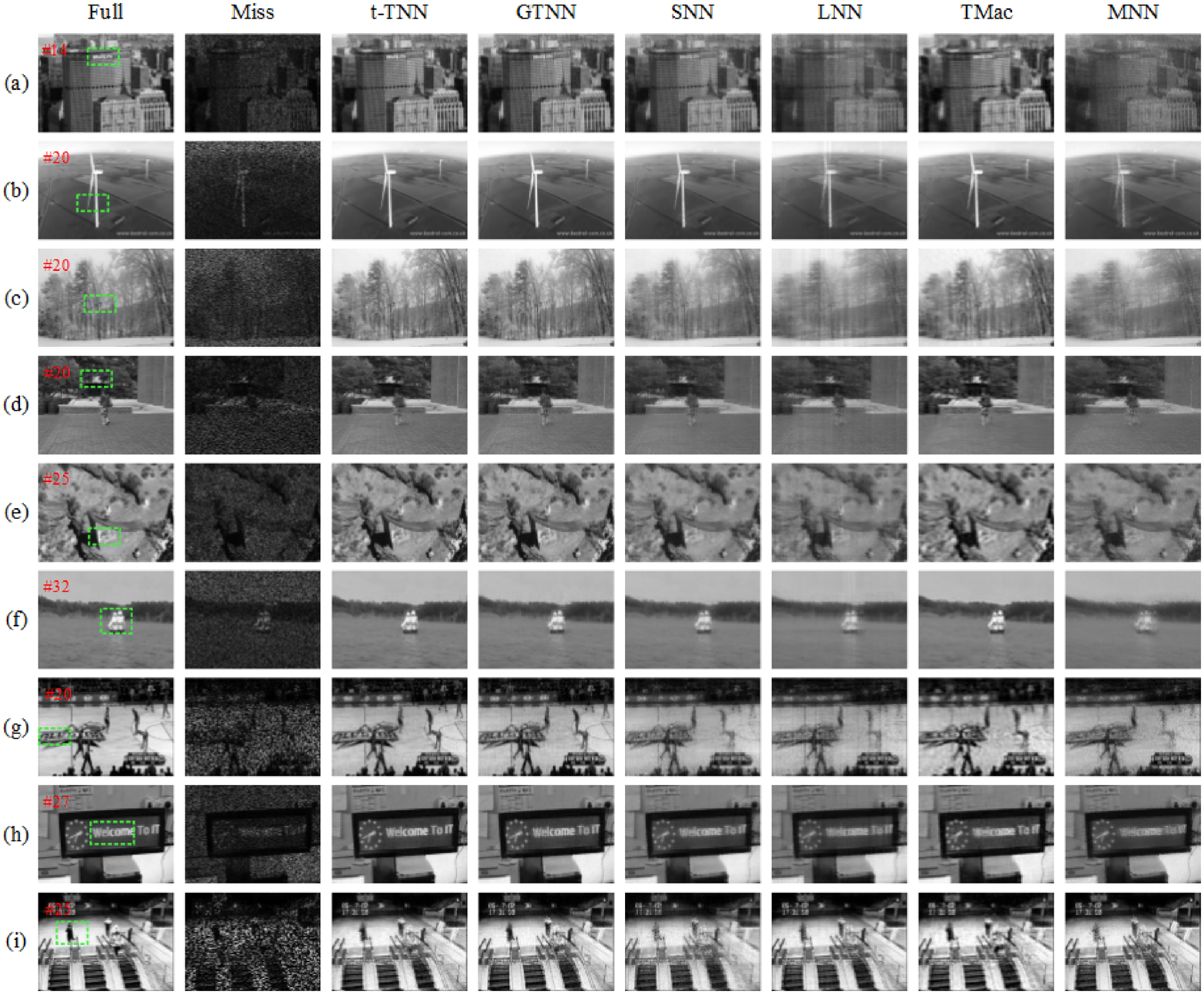}
\caption{Completion results from limited samples ($p = 0.3$). (a) \emph{Building} ($419 \times 523 \times 30$), (b) \emph{Windmill} ($304 \times 480 \times 40$), (c) \emph{Trees} ($311 \times 571 \times 37$), (d) \emph{People} ($240 \times 320 \times 40$), (e) \emph{Bike} ($288 \times 512 \times 60$), (f) \emph{Ship} ($277 \times 454 \times 60$), (g) \emph{Basketball} ($144 \times 256 \times 40$), (h) \emph{Led} ($234 \times 431 \times 60$), (i) \emph{Escalator} ($130 \times 160 \times 60$). (a)-(g) are non-stationary panning videos, whereas (h)-(i) are stationary videos.}
\label{fig:tc_results}
\end{figure*}

\begin{figure}[htb]
\setlength{\abovecaptionskip}{0pt}  
\setlength{\belowcaptionskip}{0pt} 
\renewcommand{\figurename}{Figure}
\centering
\includegraphics[width=0.48\textwidth]{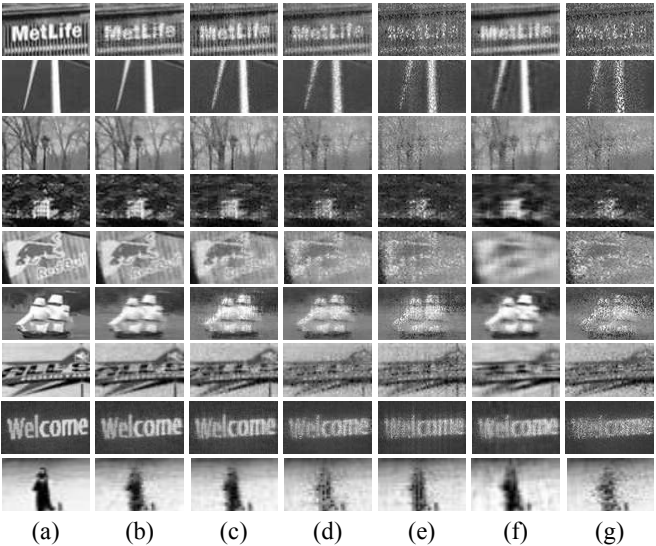}
\caption{Magnification of image content within green rectangles in Fig. \ref{fig:tc_results} with different methods. (a) Full, (b) t-TNN, (c) GTNN, (d) SNN, (e) LNN, (f) TMac, (g) MNN.}
\label{fig:tc_details}
\end{figure}

\subsection{Video completion from limited samples}\label{subsec:tc_sample}
We first test our t-TNN model by completing the video from limited entries. The entries are sampled according to the Bernoulli model, which means that each entry in the video data is sampled with probability $p$ independent of others. Fig. \ref{fig:tc_results} shows the completion result of an example frame for each video when $p = 0.3$, and the image content within green rectangle is magnified in Fig. \ref{fig:tc_details}. For videos recorded by a non-stationary panning camera [Fig. \ref{fig:tc_results}(a) - Fig. \ref{fig:tc_results}(g)], it can be clearly observed that t-TNN and GTNN, both of which exploit the spatial and temporal relationships between entries simultaneously on the t-SVD framework, perform better than other methods. For the stationary video, \emph{Led} [Fig. \ref{fig:tc_results}(h)], t-TNN and GTNN reconstruct the moving led signs more accurately, while for the stationary surveillance video, \emph{Escalator} (Fig. \ref{fig:tc_results}(i)), t-TNN and GTNN have relatively weaker advantages than other methods.

\begin{figure}[htb]
\setlength{\abovecaptionskip}{0pt}  
\setlength{\belowcaptionskip}{0pt} 
\renewcommand{\figurename}{Figure}
\centering
\includegraphics[width=0.48\textwidth]{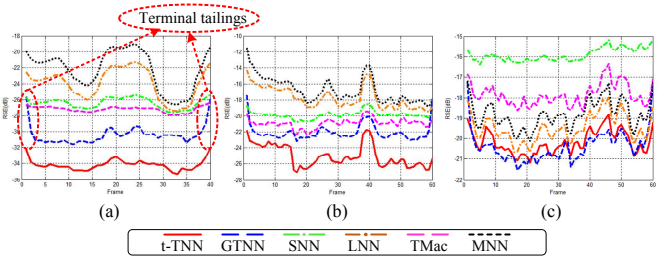}
\caption{RSE per frame (p = 0.3). For the $i$-th frame, $\mathrm{RSE}_{i} = 20\mathrm{log}_{10}  (\frac{||\mathcal{X}^{(i)} - \mathcal{M}^{(i)}||_{F}}{||\mathcal{M}^{(i)}||_{F}} )$. (a) \emph{Windmill}, (b) \emph{Led}, (c) \emph{Escalator}. }
\label{fig:tc_rse_perframe}
\end{figure}

\begin{figure}[htb]
\setlength{\abovecaptionskip}{0pt}  
\setlength{\belowcaptionskip}{0pt} 
\renewcommand{\figurename}{Figure}
\centering
\includegraphics[width=0.48\textwidth]{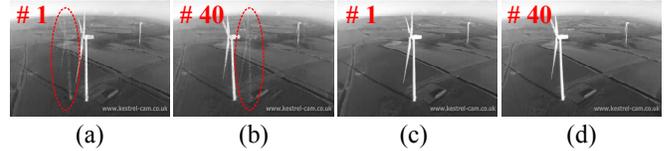}
\caption{Terminal tails (p = 0.3). (a) - (b) are terminal frames of GTNN completion result where the residual images of windmill appear. (c) - (d) are terminal frames of t-TNN completion result.}
\label{fig:tc_terminal_tails}
\end{figure}

We can also see that t-TNN recovers more textures and finer detail than GTNN in Fig. \ref{fig:tc_details}. This point is further demonstrated by the RSE curves of frames plotted in Fig. \ref{fig:tc_rse_perframe}, where three different types of videos, \emph{Windmill}, \emph{Led}, and \emph{Escalator} are taken as examples. It is worth noting that serious ``terminal tails'' exist in the GTNN completion result for the panning video \emph{Windmill} [see Fig. \ref{fig:tc_rse_perframe} (a)]. Fig. \ref{fig:tc_terminal_tails} shows the corresponding terminal frames in which the ghost of a windmill appear [see Fig. \ref{fig:tc_terminal_tails} (a) and (b)]. This phenomenon is caused by the translation of image content while the image circulant representation in the GTNN model cannot exploit the linear relationship between frames. In contrast, the t-TNN completion results effectively compress the terminal tails due to the row circulant representation over time.

Table \ref{tbl:result_rse_time} summarizes the average inverse RSE and running time for each video when $p = 0.1, 0.2, \ldots, 0.9$. The highest average iRSE and lowest average time are highlighted in bold. It can be observed that the proposed t-TNN model outperforms the other models, especially when dealing with panning videos, and a 0.8dB - 3.7dB decrease at RSE is achieved over the GTNN model. With respect to time consumption, t-TNN is significantly superior over GTNN for videos with large images.

\begin{table}[htb]
\renewcommand{\arraystretch}{1.5}
\caption{Summary of Average inverse RSE (-dB) and running time (Sec)}\label{tbl:result_rse_time}
\scriptsize
\vspace{-5pt}
\begin{center}
\setlength{\tabcolsep}{0.1cm}
$\mathrm{iRSE}_{p} = -20\mathrm{log}_{10}  (\frac{||\mathcal{X} - \mathcal{M}||_{F}}{||\mathcal{M}||_{F}} )$ for $p = 0.1~\mathrm{to}~0.9$
\begin{tabular}{||l||c|c|c|c|c|c||}
\hline\cline{1-7}
{Videos} &t-TNN                         &  GTNN              &   SNN              &  LNN        &TMac        & MNN      \\\hline\cline{1-7}
{\emph{Build.}} &\textbf{23.2} / ~\textbf{66}  &22.5 / 419          &21.2 / 153          &12.8 / 237   &18.1 / 270  &11.3 / 140 \\\hline
{\emph{W.mill}} &\textbf{38.9} / 120           &35.6 / 481          &32.1 / 102          &27.9 / 115   &28.3 / 240  &25.5 / ~\textbf{70}  \\\hline
{\emph{Trees}}  &\textbf{32.5} / 100           &28.8 / 388          &27.1 / ~\textbf{98} &21.3 / 207   &24.5 / 276  &20.7 / 107 \\\hline
{\emph{People}} &\textbf{23.1} / ~42           &19.9 / 145          &19.2 / ~52          &16.7 / ~76   &17.6 / 128  &16.4 / ~\textbf{33}  \\\hline
{\emph{Bike}}   &\textbf{28.4} / \textbf{141}  &25.2 / 408          &22.4 / 183          &16.8 / 264   &18.3 / 358  &16.2 / 156 \\\hline
{\emph{Ship}}   &\textbf{34.8} / 176           &31.7 / 653          &31.4 / 133          &25.6 / 175   &29.6 / 303  &24.0 / 112 \\\hline
{\emph{B.ball}} &\textbf{24.5} / ~30           &22.6 / ~58          &18.7 / ~25          &16.9 / ~34   &17.2 / ~63  &16.1 / ~\textbf{17}  \\\hline
{\emph{Led}}    &\textbf{29.7} / 142           &27.3 / 306          &25.0 / 129          &21.4 / 158   &23.7 / 245  &19.6 / \textbf{102} \\\hline
{\emph{Escal.}} &\textbf{23.8} / ~39           &\textbf{23.8} / ~58 &19.9 / ~11          &23.0 / ~24   &18.4 / ~53  &22.2 / ~~\textbf{9}   \\\hline\cline{1-7}
\end{tabular}
\end{center}
\end{table}

\subsection{Video completion from random occlusion} \label{subsec:tc_occlude}

\begin{figure}[htb]
\setlength{\abovecaptionskip}{0pt}  
\setlength{\belowcaptionskip}{0pt} 
\renewcommand{\figurename}{Figure}
\centering
\includegraphics[width=0.48\textwidth]{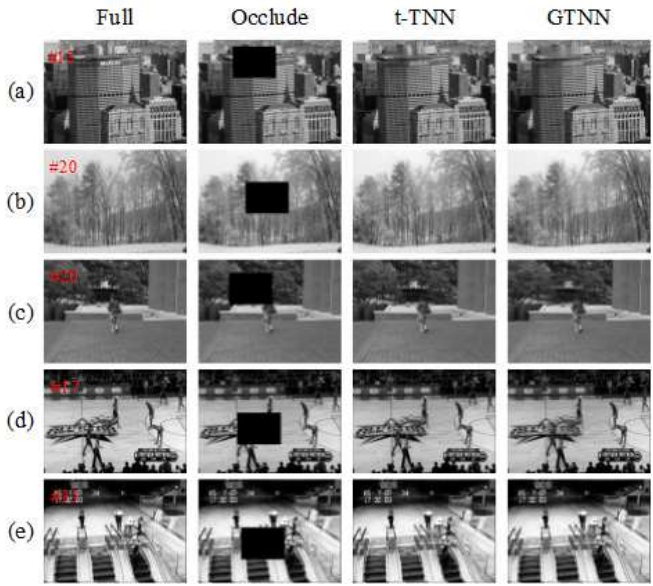}
\caption{Part completion results of t-TNN and GTNN from random occlusion. The size of the occlusion is $0.3$ times of the image size.  (a) \emph{Building}, (b) \emph{Trees}, (c) \emph{People}, (d) \emph{Basketball}, (e) \emph{Escalator}.}
\label{fig:tc_occlusion_results}
\end{figure}

\begin{figure}[htb]
\setlength{\abovecaptionskip}{0pt}  
\setlength{\belowcaptionskip}{0pt} 
\renewcommand{\figurename}{Figure}
\centering
\includegraphics[width=0.4\textwidth]{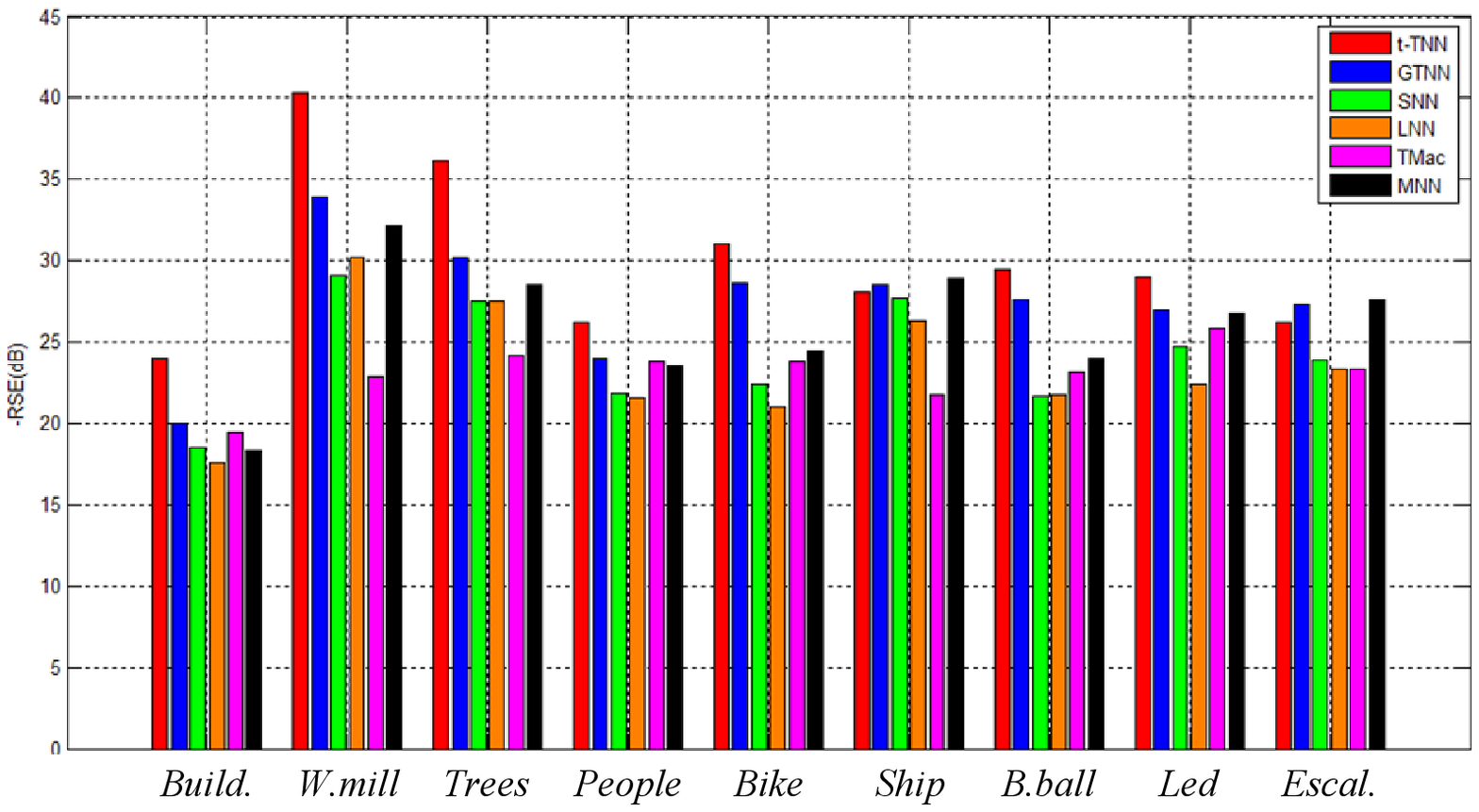}
\caption{Inverse RSE with different methods for each video. }
\label{fig:tc_occlude_rse_eachvideo}
\end{figure}

We also test the proposed t-TNN model by completing the video from the random occlusion. An image patch $0.3$ times the image size is cut out from a random position for each frame. Fig. \ref{fig:tc_occlusion_results} shows the part-completion results of t-TNN and GTNN, and Fig. \ref{fig:tc_occlude_rse_eachvideo} plots the inverse RSE for each video with different methods. Similar experiment results to those in Section \ref{subsec:tc_sample} can be observed, in which t-TNN outperforms the other methods when dealing with panning videos. More experiment results and analyses are provided in the Appendix.

\section{Conclusion}\label{sec:conclusion}
In this paper, a t-SVD based low-rank tensor model named t-TNN is proposed to complete a video from limited samples or random occlusion. The t-TNN relaxes the tensor multi-rank defined in the Fourier domain and is equal to the nuclear norm of the block circulant matricization of the twist tensor in the original domain. This two-fold action of t-TNN has two advantages, namely, efficient computation using FFT and interpretability for real problems. Furthermore, the block circulant matricization of the twist tensor can be transformed into a circulant block representation with invariance of nuclear norm, in which the entries of each circulant block correspond to mode-3 fibers of the twist tensor. In video completion, the circulant block representation compares rows of images along the time dimension in a circulant way, which exploits the temporal redundancy of videos recorded by a non-stationary panning camera and leads to superior completion performance over existing state-of-the-art low-rank models. It anticipated that the t-TNN model can be used in a wide range of applications in video processing, e.g., background modeling and video denoising.


%

%
%
%
%
%

\ifCLASSOPTIONcaptionsoff
  \newpage
\fi

\end{document}